\documentclass{article} 
\usepackage{iclr2026_conference,times}

\usepackage{amsmath,amsfonts,bm}

\def\eqref#1{equation~\ref{#1}}

\def\1{\bm{1}}

\DeclareMathAlphabet{\mathsfit}{\encodingdefault}{\sfdefault}{m}{sl}
\SetMathAlphabet{\mathsfit}{bold}{\encodingdefault}{\sfdefault}{bx}{n}

\usepackage{hyperref}
\usepackage{url}
\usepackage{graphicx}

\title{Relational Probing: LM-to-Graph Adaptation for Financial Prediction}

\author{Yingjie Niu  \\
University College Dublin\\
Belfield, Dublin, Ireland \\
\And 
Changhong Jin \\
University College Dublin \\
Belfield, Dublin, Ireland \\
\AND
Rian Dolphin \\
Massive \\
Dublin, Ireland \\
\And
Ruihai Dong \thanks{ Corresponding Author. \texttt{\{ruihai.dong\}@ucd.ie}}\\
University College Dublin \\
Belfield, Dublin, Ireland \\
\AND
}

\iclrfinalcopy 
\begin{document}

\maketitle
\vspace{-1 cm}
\begin{abstract}
    Language models can be used to identify relationships between financial entities in text.
    However, while structured output mechanisms exist, prompting-based pipelines still incur autoregressive decoding costs and decouple graph construction from downstream optimization.
    We propose \emph{Relational Probing}, which replaces the standard language-model head with a relation head that induces a relational graph directly from language-model hidden states and is trained jointly with the downstream task model for stock-trend prediction.
    This approach both learns semantic representations and preserves the strict structure of the induced relational graph.
    It enables language-model outputs to go beyond text, allowing them to be reshaped into task-specific formats for downstream models.
    To enhance reproducibility, we provide an operational definition of small language models (SLMs): models that can be fine-tuned end-to-end on a single 24GB GPU under specified batch-size and sequence-length settings.
    Experiments use Qwen3 backbones (0.6B/1.7B/4B) as upstream SLMs and compare against a co-occurrence baseline.
    Relational Probing yields consistent performance improvements at competitive inference cost.
\end{abstract}

\section{Introduction}
Stock market prediction is driven not only by a stock's own momentum but also by inter-stock relationships. Studies~\citep{pirinsky2006does, cohen2008economic, menzly2010market} indicate that related companies tend to co-move and exhibit cross-predictability.
To leverage this effect, researchers explicitly construct relational graphs based on stock relationships, including sector information~\citep{feng2019temporal, cui2023temporal}, shareholding properties~\citep{chen2018incorporating}, Wikipedia-based relevance~\citep{sawhney2021stock}, momentum spillover effects~\citep{cheng2021modeling}, and price-correlation graphs~\citep{xiang2022temporal, xia2024ci}.
However, pre-defined, static relationships (e.g., coarse industry or shareholdings) quickly become outdated in dynamic markets and can be noisy indicators of the true interaction structure. Consequently, recent studies~\citep{niu2024evaluating, niu2025ngat} construct time-varying relationship graphs from frequently updated news, rather than relying on pre-defined graphs.
In practice, news streams are disseminated at near-real-time intervals, making reliable relation extraction at scale a key challenge.

Language models have transformed text understanding and information extraction, enabling the identification of relationships between financial entities and the induction of relational graphs from news articles~\citep{chen2023chatgpt}.
However, even with structured output mechanisms, prompting-based pipelines still rely on autoregressive decoding. Not only does that result in non-trivial latency and cost, but graph construction is also typically treated as a separate stage, decoupled from downstream optimization.
For downstream financial prediction, optimizing a language model for either fluency or factual accuracy is insufficient; the predictor ultimately requires structured objects (e.g., graphs or adjacency matrices), the construction of which should be aligned with the end task.
Moreover, instruction-following benchmarks suggest that under multi-level or hierarchical constraints, LLMs remain fragile, often violating required formats or priorities.
These limitations motivate the induction of relational structure directly from language-model representations and its training jointly with the downstream predictor.

Linear probing then fine-tuning (LP$\rightarrow$FT) is advocated as a low-cost approach for learning readout heads that align general hidden states with task-specific spaces, even when the backbone is frozen or only lightly updated~\citep{howard2018universal, tomihari2024understanding}.
Linear probing (LP) first learns a near-optimal linear readout, after which fine-tuning updates features with limited, controlled drift, preserving pre-trained structure while aligning to the task.
These findings motivate a shift from unstable prompt formats~\citep{zhao2021calibrate} to stable, trainable adapters that ``pull'' representations into the target space.
We propose \emph{Relational Probing}, which replaces the standard language-modeling head with a relation head.
The relation head generates a relational graph directly from language-model hidden states and is trained jointly with the downstream model for stock-trend prediction.
This approach both learns semantic representations and preserves the strict structure of the induced relational graph.
It avoids the error-prone process of \textit{prompt $\rightarrow$ parsing structured text $\rightarrow$ post-hoc graph} and eliminates the runtime and manual effort associated with constrained decoding.

Additionally, to improve reproducibility and portability, we present an operational definition of small language models (SLMs) for end-to-end settings.
We define SLMs as models that can be jointly fine-tuned with the relation head (RH) and the downstream task model on a single 24\,GB GPU under specified batch-size and sequence-length budgets.

Our key contributions to the field include:

\begin{itemize}
    \item We generalize linear probing to relational probing, a learnable relation head converts language-model hidden states into relational graphs, aligning outputs to the target structure without constrained decoding.
    \item We jointly train the relation head to produce task-aligned, graph-structured objectives that transfer more effectively to downstream tasks.
    \item We provide an operational definition of SLMs (single-GPU, fixed-budget) and benchmark Qwen3-0.6B/1.7B/4B within a single family for controlled comparisons.
\end{itemize}

\section{Related Work}
\label{sec:related_work}

\subsection{LLMs for Relation Extraction}
Language models (LMs) can read news articles and identify relational signals (e.g., entities, event triggers, and argument roles) with minimal or no task-specific training~\citep{zhou2024grasping, wei2023chatie}. 
For relation extraction (RE), zero- or few-shot settings can be addressed via definition-conditioned prompting~\citep{zhou2024grasping} and instruction-tuned formulations~\citep{wang2023instructuie}.
At the same time, an evaluation~\citep{hamad2024fire} reports that financial RE performance remains below that of fully supervised state-of-the-art systems, yet is still practically useful.
Furthermore, recent studies show that relation extraction techniques based on language models achieve consistent improvements via prompting~\citep{wadhwa2023revisiting, li2023revisiting}, linear probing~\citep{alt2020probing}, and instruction tuning~\citep{jiao2023instruct, xu2024large}.

\subsection{Probing and Parameter-Efficient Adaptation}
Linear probing freezes the LLM and trains a lightweight head on intermediate states to predict task-specific targets~\citep{tomihari2024understanding}. 
As the head operates on continuous hidden states rather than the tokenizer/decoder, it can produce the diverse outputs required by downstream modules.
Low-Rank Adaptation (LoRA)~\citep{hu2022lora} modifies the generator itself and remains coupled to token-level decoding. Even ignoring time and memory costs, LoRA outputs are still strings (typically JSON) that must be parsed and validated before any GNN can use them.
In contrast, linear-probing outputs are ready for downstream use and can be integrated into joint training with the GAT for the FTP task.
Linear probing provides the flexibility and calibration needed to match the exact formats required by downstream tasks~\citep{tomihari2024understanding, niu2023learning}. These capabilities are beyond the scope of LoRA, which remains within the LLM text channel.

While a single linear layer is sufficient in many cases, the recent trend has been toward richer, yet still lightweight, heads when structure plays a key role. Examples include MLP heads for nonlinearity~\citep{eberts2019span} and biaffine/bilinear scorers for pairwise relations (entity--entity and token--token)~\citep{dozat2016deep, verga2018simultaneously}.
These designs retain the core advantages of probing while improving flexibility.

\subsection{Graph-based Financial Trend Prediction}
Financial trend prediction (FTP) aims to predict asset dynamics (e.g., direction, returns, or volatility) using historical signals at the stock level~\citep {rundo2019deep}. Such work typically models each stock as an isolated time series and applies sequential models such as RNNs and LSTMs~\citep{bao2017deep, fischer2018deep, sun2024research}.
However, this assumption overlooks that stock prices are influenced not only by their own history but also by related stocks via momentum spillover.
Consequently, the field has evolved from isolated asset models~\citep{bao2017deep, fischer2018deep} to relational models (e.g., GNNs) that incorporate cross-asset relationships~\citep{feng2019temporal, song2023stock, das2024integrating, wang2021review}.
These relational approaches typically follow a two-stage pipeline: (1) construct a predefined relational graph; (2) apply a downstream prediction model.
To address the limitations of static graphs in dynamic markets, research has evolved from static priors~\citep{feng2019temporal} to dynamic graphs~\citep{lei2024dr, qian2024mdgnn, liu2024multimodal}, and from price-only signals~\citep{bao2017deep, fischer2018deep} to text/event-augmented inputs~\citep{chen2023chatgpt}.

Dynamic graphs are attractive because they react to incoming information, enabling same-day relational reconstruction~\citep{schwenkler2020network}.
Building on this trajectory, we integrate text understanding, graph construction, and prediction into a single loop via relational probing, using downstream FTP performance as a benchmark for evaluation.

\section{Relational Probing}
\label{sec:relation_probing}

\begin{figure}[h]
    \centering
    \includegraphics[width=\linewidth]{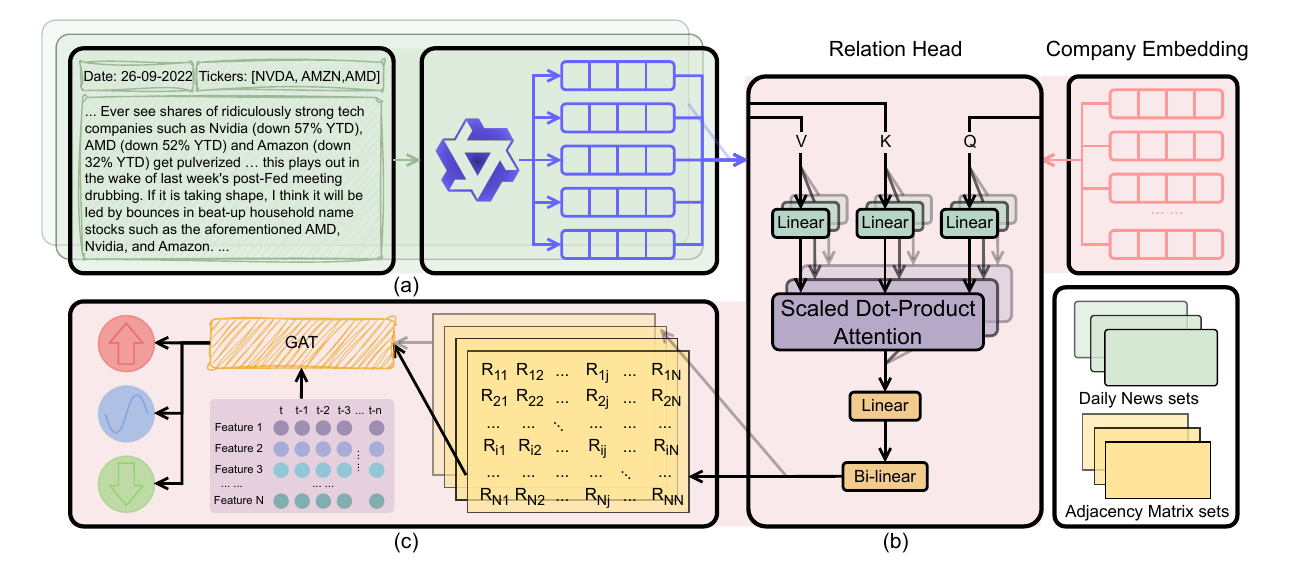}
    \caption{End-to-End Language Model → Relational Probing → GAT framework for Financial Trend Prediction: (a) News encoding; (b) A lightweight Relation Head (RH) maps language-model hidden states to adjacency matrices; (c) The GAT uses the dynamic graphs with node features to predict next-step trends.}
    \label{fig:framework}
\end{figure}

This section presents \emph{Relational Probing}, an end-to-end framework that couples an upstream language-model backbone with a downstream graph attention network (GAT)~\citep{velivckovic2017graph} for financial trend prediction.
Given a selection of news articles, the method constructs a day-level ticker graph and uses it to predict stock trends via a downstream graph attention network (GAT)~\citep{velivckovic2017graph}.
Figure~\ref{fig:framework} summarizes the pipeline:
\begin{itemize}
    \item \textbf{(a) News Encoding.} Each news article is tokenized and encoded by an LM to produce contextual token representations.
    \item \textbf{(b) Relation Induction.} A lightweight Relation Head (RH) maps language-model hidden states to adjacency matrices. These signals are aggregated across all news articles on a given day to form a structured relational graph at the day level.
    \item \textbf{(c) Downstream Prediction.} The GAT uses the induced graph (adjacency and edge weights) and the corresponding node features to predict the next-step trend for each ticker.
\end{itemize}

Importantly, the RH and GAT components are trained jointly: gradients from the downstream prediction loss backpropagate through the GAT into the RH, aligning relation induction with the end financial prediction goal.

\subsection{Relation head: from news articles to interaction matrix}
Given a news article, let $P = \{p_0, p_1, \ldots, p_{L-1}\}$ denote the prompted token sequence of length $L$.
A language-model backbone $M$ encodes $P$ and produces contextual token representations
\begin{equation}
    H = M(P) = \{h_0, h_1, \ldots, h_{L-1}\} \in \mathbb{R}^{L \times d},
\end{equation}
where $h_i \in \mathbb{R}^d$ is the hidden state of the final-layer at position $i$ and $d$ is the hidden size.

Let $\mathcal{V} = \{1, \ldots, N\}$ denote the ticker set.
We initialize learnable ticker embeddings $E = \{e_1, \ldots, e_N\} \in \mathbb{R}^{N \times d}$.
To condition tickers on the news article content, each ticker embedding attends to token states via scaled dot-product attention.
Define $Q = EW_Q$, $K = HW_K$, and $V = HW_V$, with $W_Q, W_K, W_V \in \mathbb{R}^{d \times d}$.
The news-conditioned ticker embeddings are
\begin{equation}
    E' = \mathrm{Softmax}\left(\frac{QK^T}{\sqrt{d_k}}\right)V \in \mathbb{R}^{N \times d},
    \label{equ:E'}
\end{equation}
where $d_k$ is the key dimension (we set $d_q = d_k = d_v = d$ in the experiment).
We then reduce dimensionality and stabilize optimization by a linear projection followed by layer normalization:
\begin{equation}
    Z = \mathrm{LayerNorm}(E'W_{\mathrm{proj}}), \quad W_{\mathrm{proj}} \in \mathbb{R}^{d \times d_p}, \quad Z \in \mathbb{R}^{N \times d_p}.
    \label{equ:Z}
\end{equation}
We use $Z$ as a news-conditioned ticker representation for relation induction (i.e., to compute ticker--ticker interaction scores). 
In our framework, the node features used by the downstream GAT are encoded from historical trading data (see Section~\ref{sec:gat_pred}).
Pairwise interactions between tickers are captured with a bilinear map in the reduced space, again followed by normalization:
\begin{align}
    I &= Z W_{\mathrm{bi}} Z^T, \quad W_{\mathrm{bi}} \in \mathbb{R}^{d_p \times d_p}, \quad I \in \mathbb{R}^{N \times N}.
    \label{equ:I} \\
    \tilde{I} &= \mathrm{LayerNorm}(I).
    \label{equ:I_tilde}
\end{align}

\subsection{Daily graph construction}
Graph construction is performed at the daily level. Let $D_i^t$ be the $i$-th article on day $t$, and let $n_t$ denote the number of articles on that day.
Applying Eqs.~\ref{equ:E'}--\ref{equ:I_tilde} to each $D_i^t$.
We aggregate and normalize news-level evidence into day-level edge attributes:
\begin{equation}
    attr^t = \mathrm{LayerNorm}\left(\sum_{i=1}^{n_t} \tilde{I}_i^t\right) \in \mathbb{R}^{N \times N}.
    \label{equ:attr}
\end{equation}
To obtain a sparse adjacency, we apply a thresholding operator $\tau$ (optionally zeroing the diagonal to remove self-loops):
\begin{equation}
    A^t_{uv} = \mathbb{I}\left[attr^t_{uv} > \tau\right], \quad A^t \in \{0,1\}^{N \times N}.
    \label{equ:adj}
\end{equation}
We treat $attr^t$ as a real-valued edge attribute matrix and use the masked edge weights
\begin{equation}
    W^t = A^t \odot attr^t.
    \label{equ:edge_weight}
\end{equation}
This yields a day-level weighted graph $\mathcal{G}^t = (\mathcal{V}, \mathcal{E}^t, W^t)$, where $(u,v) \in \mathcal{E}^t$ iff $A^t_{uv}=1$.

\subsection{Downstream GAT and prediction}
\label{sec:gat_pred}
The downstream GAT operates on node features encoded from historical trading data.
For each ticker $u$ on day $t$, let $S_u^t \in \mathbb{R}^{T \times p}$ denote its lookback window of $T$ trading steps with $p$ market features (e.g., price/volume-derived signals).
We encode $S_u^t$ using an LSTM to obtain the day-level node feature $x_u^t \in \mathbb{R}^{d_p}$:
\begin{equation}
    x_u^t = \mathrm{LSTM}(S_u^t), \quad X^t = \{x_1^t, \ldots, x_N^t\} \in \mathbb{R}^{N \times d_p}.
    \label{equ:lstm_node}
\end{equation}
Given $\mathcal{G}^t$, the GAT takes $H^{(0)} = X^t$ as input.
For each layer $\ell$, we compute the attention coefficients on the neighbors and update node representations:
\begin{align}
    e_{uv}^{(\ell)} &= \mathrm{LeakyReLU}\left(\mathbf{a}^T\left[W^{(\ell)}h_u^{(\ell)}\,\Vert\,W^{(\ell)}h_v^{(\ell)}\right]\right),
    \label{equ:gat_e}\\
    \alpha_{uv}^{(\ell)} &= \mathrm{softmax}_{v \in \mathcal{N}(u)}\left(e_{uv}^{(\ell)}\right),
    \label{equ:gat_alpha}\\
    h_u^{(\ell+1)} &= \sigma\left(\sum_{v \in \mathcal{N}(u)} \alpha_{uv}^{(\ell)}\, W^t_{uv}\, h_v^{(\ell)}\right),
    \label{equ:gat_update}
\end{align}
where $\mathcal{N}(u)=\{v: A^t_{uv}=1\}$, $\Vert$ denotes concatenation, and $\sigma$ is a nonlinearity.
We then predict the trend label for each ticker using a linear classifier:
\begin{equation}
    \hat{y}_u^t = \mathrm{softmax}(W_o h_u^{(L)} + b_o).
    \label{equ:pred}
\end{equation}

\subsection{Joint training objective}
Let $y_u^t$ denote the ground-truth label for ticker $u$ on day $t$.
We optimize the cross-entropy loss over labeled tickers and days:
\begin{equation}
    \mathcal{L} = - \sum_t \sum_{u \in \mathcal{V}_t} \log \hat{y}_u^t[y_u^t],
    \label{equ:loss}
\end{equation}
where $\mathcal{V}_t$ is the set of labeled tickers on day $t$.
Gradients from $\mathcal{L}$ backpropagate through the GAT and the relation head into the language model, aligning relation induction with the downstream financial objective.

\section{Results \& Discussion}
\subsection{Implementation Details}
We use the financial news dataset with complete ticker labels~\citep{niu2024evaluating}, which contains stock trading data and corresponding news articles to build dynamic financial relational graphs. 
We follow the original study in terms of dataset splitting. The training, validation, and test sets are split in an 8:1:1 ratio. We select the graph attention network (GAT) as the downstream GNN architecture and use class-weighted cross-entropy loss during training, both of which are kept exactly the same as the original study. 
Experiments are implemented in PyTorch and PyTorch Geometric with a fixed random seed of $1.3423$. We set the projection dimension $d_p \in \{128, 256\}$ and the threshold $\tau = 0.5$.

Following our definition of small language models (SLMs), all components of the framework, including the SLM, the RH, and the GAT, are jointly trained under predefined budgets for batch size and sequence length.
All experiments are conducted on an NVIDIA GeForce RTX 4090 GPU using the Adam optimizer~\citep{adam2014method}. We sweep the learning rate $\mathrm{lr} \in \{10^{-3}, 10^{-4}, 10^{-5}\}$. We train for up to $30$ epochs with early stopping.

For the SLM backbone, we adopt the Qwen3 series, which is a strong baseline in this parameter regime~\citep{yang2025qwen3}.
To minimize variability across model families, we select models from the same architecture, which are Qwen3-0.6B, Qwen3-1.7B, and Qwen3-4B, to avoid confounding effects from architectural differences.
We employ a single fixed prompt template to extract relations from news articles  (Figure~\ref{fig:prompt-news}).

\begin{figure}[hb]
\centering
\fbox{%
\parbox{0.85\columnwidth}{%
Please read the following news text and determine whether there is a connection between the stocks \{\textit{TICKERS}\} and whether they may influence each other.
\\ Hint: There may be positive or negative correlations.
\\ News Article: \{\textit{NEWS\_TEXT}\}
}
}
\caption{Prompt used to extract relations from news.}
\label{fig:prompt-news}
\end{figure}

Consistent with previous studies~\citep{hu2018listening, prachyachuwong2021stock, niu2024evaluating}, we frame the stock trends prediction as a three-class task (down, up, or unchanged). Any rate of return beyond the absolute value of the standard deviation will be regarded as a true up/down trend. For the node $u$ on day $t$, the label $y_u^t$ is 
\begin{align}
y_u^t= 
\begin{cases}
    negative,  & \text{if $r_u^{t+1} < -\mathrm{std}(r_u)$} \\
    neutral,  & \text{if $-\mathrm{std}(r_u) < r_u^{t+1} < \mathrm{std}(r_u)$} \\
    positive, & \text{if $r_u^{t+1} > \mathrm{std}(r_u)$}
\end{cases}
\end{align}%
where $r_u^{t+1}$ is the rate of return of node $u$ on day $t+1$, $r_u$ represents the time series of returns for node $u$ over the full dataset period, and $\mathrm{std}(r_u)$ is the standard deviation of $r_u$. 
Using three classes is beneficial from an investment management perspective. Investors typically focus on firms with large positive or negative returns, as these have greater profit potential, while small fluctuations are considered normal.
The distribution of three classes in our dataset is shown in Table \ref{tab:class_distribution}.

\begin{table}[h]
\caption{Classes distribution in train/validation/test sets.}
\label{tab:class_distribution}
\centering
\renewcommand{\arraystretch}{1.2} 
\begin{tabular}{lccc}
\hline
Dataset        & Positive & Neutral & Negative \\
\hline
Train set      & 0.1725 & 0.6742 & 0.1533 \\
Validation set & 0.1091 & 0.7401 & 0.1507 \\
Test set       & 0.1440 & 0.6777 & 0.1783 \\
\hline
\end{tabular}
\end{table}


\subsection{Overall Performance}

We compare relational graphs constructed using our proposed relational probing with graphs constructed using a baseline method based on news co-occurrence. Each graph is then used as input to the downstream task of predicting stock trends. 
To ensure a fair comparison, we keep the GAT architecture, optimization settings, and data split fixed across all experiments; therefore, any performance differences can be attributed solely to the choice of relational graphs. 

\begin{table}[h]
\caption{Comparison of GAT performance using different graphs (five independent runs).}
\label{tab:graphs_comparison}
\centering
\renewcommand{\arraystretch}{1.2} 
\begin{tabular}{lcccc}
\hline
Graph Construction & Accuracy & Macro F1 & MCC & AUC \\
\hline
Majority-only & \textbf{0.6777} & 0.2693 & 0.0000 & 0.5000 \\
Co-occurrence & 0.4471 & 0.2831 & 0.0143 & 0.5232 \\
Qwen3-0.6B    & 0.5339 & 0.3171 & 0.0312 & 0.5488 \\
Qwen3-1.7B    & 0.5664 & 0.3221 & 0.0435 & 0.5513 \\
Qwen3-4B      & 0.5705 & \textbf{0.3272} & \textbf{0.0562} & \textbf{0.5571} \\
\hline
\end{tabular}
\end{table}

Table~\ref{tab:graphs_comparison} reports the mean performance over five independent runs using four metrics: accuracy, macro F1, Matthews correlation coefficient (MCC)~\citep{chicco2020advantages}, and AUC~\citep{fawcett2006introduction}.
To contextualize the impact of label imbalance, we additionally include a majority-only baseline that predicts the majority label (neutral) for every sample. This trivial classifier attains the highest accuracy (0.6777), yet yields a macro F1 of 0.2693, an MCC of 0.0000, and an AUC of 0.5000, indicating no discriminative power beyond the prior label distribution. 
This result highlights that, in our setting, accuracy is skewed by the class imbalance and can be misleading as a primary indicator of progress. In contrast, for real-world stock-trend prediction, correctly identifying movements in the tails of the returns distribution (up/down classes) is typically more valuable than labeling neutral days; this motivates evaluating with macro F1, MCC, and AUC and placing greater emphasis on these imbalance-aware metrics.

In this context, we make two \textbf{key observations}. 
Firstly, graphs induced by relational probing perform consistently better than the co-occurrence baseline across all metrics. In particular, Qwen3-0.6B improves macro F1 from 0.2831 to 0.3171 (+0.0340) and MCC from 0.0143 to 0.0312 (+0.0169), with corresponding improvements in accuracy and AUC. 
These results suggest that simple co-occurrence edges introduce substantial noise (e.g., spurious co-mentions) and fail to capture the context-dependent semantics required for meaningful relational induction. In contrast, relational probing extracts semantically grounded interactions from language-model hidden states, producing a cleaner, more informative relational structure for message passing.

Secondly, performance scales with SLM capacity, albeit with diminishing returns. 
Increasing the SLM size further improves downstream results: Qwen3-1.7B reaches an accuracy of 0.5664 and a macro F1 of 0.3221, while Qwen3-4B achieves the best overall scores (accuracy of 0.5705, macro F1 of 0.3272, MCC of 0.0562, AUC of 0.5571). Notably, the incremental gain from 1.7B to 4B is smaller than that from co-occurrence to 0.6B, suggesting diminishing returns once the dominant semantic relations are already captured. This trend is consistent with larger SLMs producing more accurate token representations and more discriminative relation scores, resulting in more reliable edges and edge weights.

Furthermore, we observe that the MCC value remains small, with a maximum of 0.0562, and that the AUC is at most 0.5571. This behavior is to be expected in a class-imbalanced and noisy stock-trend setting, where even meaningful improvements in minority detection and ranking can result in minimal absolute changes in metrics.
It is important to note that the observed improvements are not a by-product of changing the downstream learning objective; both the co-occurrence baseline and our relational-probing graphs use the same downstream GAT architecture, the same data split, and the same class-weighted cross-entropy loss (with minority classes weighted at five times the weight of the majority class). This combination of loss function and class weighting explicitly discourages the majority-only strategy and aligns training with minority sensitivity.
Under this controlled setup, relational probing cannot ``solve'' class imbalance by itself. Rather, it provides complementary value by improving the quality of relational signals, reducing co-occurrence noise, and inducing semantically meaningful edges.

\section{Ablation Study}
We conducted an ablation study on the relation head (RH) architecture. Specifically, a comparison is performed between three different head constructions:

\begin{itemize}
    \item \textit{Pooling}: Apply average pooling on $H$ as follows $h_{avg}= \frac{1}{L-1}\sum_{i=0}^{L-1}h_i, h_i\in H $. Then $I$ is calculated by the dot-product of two different linear transformations of $h_{avg}$. 
    \item \textit{Limited}: Restricting the range of cross-attention only between news and its mentioned tickers. Formally, we replace the query Q in Equation \ref{equ:E'} by $\tilde{Q}=\tilde{E}W_q$, where $\tilde{E}=\{e_i | i\in X\}$, and repeat Equation \ref{equ:E'} to \ref{equ:attr}.
\end{itemize}
The ablation study focuses on the comparison between our proposed full-range attention RH (hereinafter referred to as ``\textit{Full}'') with ``\textit{Limited}'' and ``\textit{Pooling}'' construction strategies. 

\begin{table}[h]
\caption{Different architectures of relation head.}
\label{tab:heads_comparison}
\centering
\renewcommand{\arraystretch}{1.2} 
  \begin{tabular}{llcccc}
    \hline
    SLM & Head & Accuracy & Macro F1 & MCC & AUC\\
    \hline
    Qwen3-0.6B & Pooling & 0.2542 & 0.1798 & 0.0105 & 0.5490 \\
    Qwen3-0.6B & Limited & 0.5398 & 0.3066 & 0.0142 & 0.5107 \\
    Qwen3-0.6B & Full & 0.5339 & 0.3171 & 0.0312 & 0.5488 \\
    \hline
    Qwen3-1.7B & Pooling & 0.3326 & 0.2352 & 0.0061 & 0.5457 \\
    Qwen3-1.7B & Limited & 0.5540 & 0.3050 & 0.0038 & 0.5071 \\
    Qwen3-1.7B & Full & 0.5664 & 0.3221 & 0.0435 & 0.5513 \\
    \hline
    Qwen3-4B & Pooling & 0.3544 & 0.2458 & 0.0071 & 0.5482 \\
    Qwen3-4B & Limited & 0.5252 & 0.3022 & 0.0129 & 0.5124 \\
    Qwen3-4B & Full & 0.5705 & 0.3272 & 0.0562 & 0.5571 \\
  \hline
  \end{tabular}
\end{table}

\begin{table}[h]
\caption{Different context of hidden states to relation head.}
\label{tab:hidden_comparison}
\centering
\renewcommand{\arraystretch}{1.2} 
  \begin{tabular}{lccccc}
    \hline
    SLM & Context & Accuracy & Macro F1 & MCC & AUC\\
    \hline
    Qwen3-0.6B & IO & 0.5339 & 0.3171 & 0.0312 & 0.5488 \\
    Qwen3-0.6B & I+G & 0.5488 & 0.3204 & 0.0256 & 0.5467 \\
    \hline
    Qwen3-1.7B & IO & 0.5664 & 0.3221 & 0.0435 & 0.5513 \\
    Qwen3-1.7B & I+G & 0.4935 & 0.3008 & 0.0173 & 0.5449 \\
    \hline
    Qwen3-4B & IO & 0.5705 & 0.3272 & 0.0562 & 0.5571 \\
    Qwen3-4B & I+G & 0.5504 & 0.3247 & 0.0410 & 0.5600 \\
    \hline
  \end{tabular}
\end{table}

\subsection{Theoretical Derivation} 
To simplify the derivation, we assume that GAT is completely effective in an ideal state, that is, when the input graph information fully reflects the real information, GAT can predict the real stock fluctuations, i.e., $Y^{t+1} = GAT(V^t, R^t)$. Given that the difference between graph construction methods lies in the connections rather than the node representation, we assume here that $V^t$ is the optimal node representation. 
The objective of this study is to construct a graph connection $attr^t \to R^t$. Ideally, $R^t$ should be constructed based on the complete information set $F^t$ available at time $t$. This would include transaction data, social media, and internal company information, among other sources. However, $F^t$ is obviously unavailable. In our experiment, $attr^t$ is constructed based on a partial news set $X^t$ from day $t$. Since $X^t \ll F^t$, the ``Limited'' constructed $attr^t_{limited}$ is clearly far from the true $R^t$. The ``Full'' constructed $attr^t_{full}$ is able to compensate for the lack of information sources in $X^t$ to some extent. 
Therefore, the distance between $attr^t_{full}$ and $R^t$ is closer than that between $attr^t_{limited}$ and $R^t$, i.e. $Dist(attr^t_{full}, R^t) < Dist(attr^t_{limited}, R^t)$. 

\subsection{Empirical Proof} 
The results in Table \ref{tab:heads_comparison} show that both ``\textit{Limited}'' and ``\textit{Full}'' attention RH outperform the \textit{Pooling} construction. This demonstrates that establishing relationships through semantic fusion requires more than simply pooling averages, providing a robust demonstration of the effectiveness and necessity of the attention RH that was designed. 
Moreover, the performance of ``\textit{Full}'' is shown to perform at a higher level than ``\textit{Limited}'' attention RH, which is consistent with our derivation.

In addition to the exploration of varying RH architectures, we also investigate the effect of using different LLM backbone hidden states on the final performance. Specifically, under the setting where all models employ the ``\textit{Full}'' attention RH, the comparison is between two variants: 
1) using only the hidden states of the input sequence (denoted as \textit{Input-only, IO}); and 
2) using the hidden states from input and the model-generated tokens (denoted as \textit{Input+Gen, I+G}). 
As shown in Table~\ref{tab:hidden_comparison}, \textit{Input+Gen} yields performance similar to or less effective than \textit{Input-only}. We therefore conclude that incorporating model-generated tokens during graph construction does not provide significant performance gains. Furthermore, given that the process of inferring on large-scale news text substantially increases computational overhead (in our experiments, \textit{Input+Gen} was tens of times slower than \textit{Input-only}), it can be argued that the addition of generated tokens is unnecessary in practice.

\section{Conclusion}
We introduced \emph{relational probing}, an end-to-end framework that couples an upstream language model with a downstream graph attention network (GAT) for financial trend prediction (FTP).
The standard language-model head was replaced by a lightweight relation head, which inducing a relational graph directly from hidden states and is trained jointly with the downstream predictor.
This design avoids autoregressive decoding overhead, reduces pipeline complexity, and aligns induced graph with the downstream task.
Across five independent runs with a fixed downstream architecture and data split, relational probing produced higher-quality graphs that aligned better with downstream predictors than the co-occurrence baseline. 
Moreover, we found that increasing the capacity of small language models improved graph quality and downstream metrics, albeit with diminishing returns. These findings suggest that relational probing benefits from stronger representations while remaining practical at relatively small scales. To promote reproducibility, we provide an operational definition of small language models and report results under standardized configurations.
In future work, we will (i) investigate principled ways to handle label imbalance (e.g., calibrated losses or re-weighting) without changing the underlying data split, in order to quantify gains that are complementary to relational probing, and (ii) explore a family of task-specific heads that can be swapped without altering the backbone while maintaining structured outputs (e.g., sparse weighted graphs).


\subsubsection*{Acknowledgments}
This publication has emanated from research conducted with the financial support of Taighde Éireann - Research Ireland through the Research Ireland Centre for Research Training in Machine Learning (Grant No. 18/CRT/6183). For the purpose of Open Access, the author has applied a CC BY public copyright licence to any Author Accepted Manuscript version arising from this submission.

\bibliography{iclr2026_conference}
\bibliographystyle{iclr2026_conference}

\appendix
\section{The Use of Large Language Models (LLMs)}
We utilized a large language model (LLM) as a general-purpose assist tool to aid or polish
the writing. The LLM’s role was strictly limited to improving grammar, spelling, and clarity. 
It was not used for research ideation, experiment design, data analysis, code, or generating scientific content. 
The authors take full responsibility for the content, originality, and accuracy of the paper.

\end{document}